\def\year{2018}\relax
\documentclass[letterpaper]{article} %DO NOT CHANGE THIS
\usepackage{aaai18}  %Required
\usepackage{times}  %Required
\usepackage{helvet}  %Required
\usepackage{courier}  %Required
\usepackage{url}  %Required
\usepackage{graphicx}  %Required
\usepackage{comment}
\usepackage{amsmath,bm}
\usepackage{bbm,color}
\usepackage{multirow}
\usepackage{tabularx}
\usepackage{amssymb}

\usepackage[T1]{fontenc}
\usepackage{aecompl}

\begin{document}
% The file aaai.sty is the style file for AAAI Press 
% proceedings, working notes, and technical reports.
 \renewcommand\footnotemark{}
\title{Learning the Joint Representation of Heterogeneous Temporal Events 
\\for Clinical Endpoint Prediction}
%\author{%Luchen Liu\\
%Association for the Advancement of Artificial Intelligence\\
%2275 East Bayshore Road, Suite 160\\
%Palo Alto, California 94303\\
%}
\author{Luchen Liu\textsuperscript{1}, Jianhao Shen\textsuperscript{1}, Ming Zhang\textsuperscript{1*}\thanks{*corresponding authors}, Zichang Wang\textsuperscript{1}, Jian Tang\textsuperscript{2,3*} \\
\textsuperscript{1}School of EECS, Peking University, Beijing China \\
\textsuperscript{2}HEC Montreal, \textsuperscript{3}Montreal Institute for Learning Algorithms \\
liuluchen@pku.edu.cn,
jhshen@pku.edu.cn,
mzhang\_cs@pku.edu.cn,
dywzc123@163.com,
tangjianpku@gmail.com
}
% \textsuperscript{1}
% \And Jianhao Shen\textsuperscript{1} \\ \textsuperscript{1}Department of Electrical Engineering and Computer Science \\Peking University \\ Beijing, China
% \And Ming Zhang\textsuperscript{1}\thanks{corresponding author} \And Zichang Wang\textsuperscript{1} \\\textsuperscript{2} Montreal Institute for Learning Algorithms\\ University of Montreal \\ Montreal, Canada
% \And Jian Tang\textsuperscript{2}}
\maketitle
\begin{abstract}
%important
%Clinical endpoint prediction (e.g., death prediction)  plays an important role in providing a better quality of intelligent health care.  The most important resource to help achieve accurate clinical endpoint  prediction is the massive multiple sequential data in (EHR).
The availability of a large amount of electronic health records (EHR) provides huge opportunities to improve health care service by mining these data. One important application is clinical endpoint prediction, which aims to predict whether a disease, a symptom or an abnormal lab test will happen in the future according to patients' history records. This paper develops deep learning techniques for clinical endpoint prediction, which are  effective in many practical applications. However, the problem is very challenging since patients' history records contain multiple heterogeneous temporal events such as lab tests, diagnosis, and drug administrations. The visiting patterns of different types of events vary significantly, and there exist complex nonlinear relationships between different events. In this paper, we propose a novel model for learning the joint representation of heterogeneous temporal events. 
The  model adds a new gate to control the visiting rates of different events which effectively models the irregular patterns of different events and their nonlinear correlations. 
Experiment results with real-world clinical data on the tasks of predicting death and abnormal lab tests prove the effectiveness of our proposed approach over competitive baselines. 

%The problem is very challenging since 

%This paper studies developing deep learning techniques for 
% However, modeling EHR data for clinical prediction is challenging, because it is difficult to model the temporal dependency of thousands of clinical sequences when making predictions. And the records in different type of clinical sequences is sparsely  logged  with asynchronous multi-scale periods.
%  %our novelty...
% To address these challenges, we propose Multi-sequence

% LSTM to model patients’ multiple sequential health records. Not only the content of patients' record but also their sequence types are embedded into related semantic spaces. And we equip the following RNN with sequence gates in order to guide 
% neurons in the hidden layer to focus on different clusters of  correlated sequences in different rhythm. 
% %Results
% Experimental results on real clinical data show that our method accurately predicts clinical endpoint outcomes such as death and abnormal lab test results.
% The joint representations of multiple sequences are effective to explore the temporal dependency of multiple sequential data and achieve significantly improved performance on real datasets for clinical endpoint prediction tasks.

\end{abstract}

\section{Introduction}

%problem: multi-sensor clinical data from multiple source 
The volume of electronic health records (EHR) is expanding at a staggering rate, providing a great opportunity for machine learning and data mining researchers to analyze these data so as to provide better health care service. An important application of machine learning in health care is predicting the clinical endpoints such as a disease, symptom, or laboratory abnormality based on patients' historical records.  
%In clinical research, a clinical endpoint generally refers to occurrence of a disease, symptom, sign or laboratory abnormality that constitutes one of the target outcomes.

This paper develops effective deep learning techniques for clinical endpoint prediction since deep learning techniques have been proved effective for predictive analysis in a variety of applications such as image recognition~\cite{he2016deep}, speech recognition~\cite{hinton2012deep}, and natural language understanding~\cite{blunsom2017characters}. The goal of deep learning is to learn effective semantic representations of the high-dimensional data such as images, speeches and natural language. Therefore, our goal is to effectively represent patients' historical records.

\begin{figure}[!t]
\centering
\includegraphics[width=.8\linewidth]{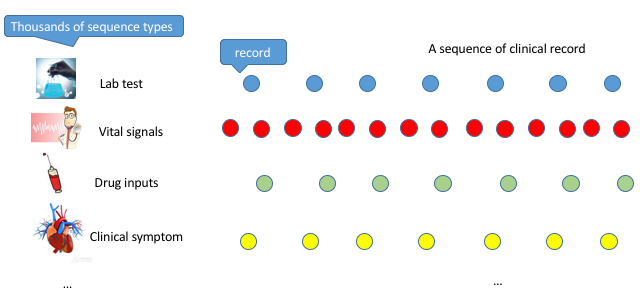}

\caption{Heterogeneous Temporal Events. The sampling rates of different events vary significantly from each other. Different kinds of events are highly correlated.}

\label{fig:distr time interval}
\end{figure}

However, the problem is challenging since patients' historical records contain a variety of heterogeneous temporal events such as different lab tests, routine vital signals, diagnosis, and drug administrations (See Fig.~\ref{fig:distr time interval} as an example). The visiting rates of different events vary significantly. For example, a patient may take a blood test every morning while take a temperature test every two hours. Besides, there is a high level of dependency among different kinds of events. For instance, some diagnosis are made according to the results of some lab tests. As a result, these heterogeneous temporal events yield heterogeneous event sequences consisting of thousands of correlated event types, the visiting rate of which varies significantly. 

In the literature, learning representations of sequences are widely studied especially in the domain of speech recognition and natural language understanding. The state-of-the-art approaches for sequence modeling are recurrent neural networks~\cite{mikolov2010recurrent} (RNNs) with the Long Short-term Memory (LSTM) units~\cite{hochreiter1997long}. RNNs are commonly used for modeling homogeneous sequences, but it is nontrivial to apply them for modeling heterogeneous event sequences. There are some recent works based on multi-task Gauss Process (MTGP)~\cite{ghassemi2015multivariate} for modeling the correlations between multiple sequences. However, the computational cost of MTGP is too expensive for EHR data since there are thousands of types of events. Therefore, we are seeking an approach that is able to: (1) effectively model the irregular visiting patterns of different events; (2) model the complex nonlinear relationships between different events; (3) scale up a large number of different types of events. 

In this paper, we propose such an approach called Heterogeneous Event LSTM(HE-LSTM) for learning the joint representation of heterogeneous event sequences. Our approach is an extension of Phased LSTM~\cite{neil2016phased}, which was recently proposed and is used to model irregular event-based sequential data. Compared to the vanilla LSTM model, Phased LSTM~\cite{neil2016phased}  adds a new time gate, which is able to naturally integrate inputs from several sensors of arbitrary sampling rates. But Phased LSTM is not suitable for modeling the heterogeneous event sequence with thousands of event types in longitudinal EHR data. Our proposed model extends it by modeling correlated heterogeneous events with multi-scale sampling rates. 
Each event type and its attributes are embedded and fed into HE-LSTM. The HE-LSTM is equipped with an event gate controlled by the event type embeddings and the their timestamps.
With the help of the event gates, the HE-LSTM can perfectly trace the temporal information of different event types in the long heterogeneous event sequence
by asynchronously sample important and related events in the heterogeneous event sequence. 
Therefore, the representation of heterogeneous temporal events can be updated base on the dependency of the current input event and other events maintained in the HE-LSTM.
%To address the above challenges, we propose Multi-sequence LSTM to learn the joint representation of 
% patients’ multiple sequential health records for the endpoint prediction.

% the multi-scale period of clinical sequences
% the temporal Dependency of related clinical sequences

% In this model,the time stamp is the input of the time gate which controls the update of the cell state, the hidden state and
% thus the final output. Meanwhile, only samples lying in the
% model’s active state are utilized, resulting in sparse updates
% during training. However, there
% exist several challenges preventing Phased LSTM from becoming
% the best fit for clinical endpoint prediction tasks from multiple sequential data.

% First of all, phased LSTM cannot automatically learn multi-scale period pattern from different clinical sensors. Secondly, phased LSTM
% cannot  explicitly learn the temporal dependency of correlated sequences.

%%% our proposed approach 

We conduct extensive experiments on real-world clinical data. Experiment results on the tasks of death prediction and abnormal lab test prediction prove that our proposed approach outperforms competitive baselines. Our proposed approach can be widely used in modeling data collected from sensors with arbitrary sampling rates, such as data collected from mobile sensors. 

Our main contributions are:

%3 contributions
$\bullet$ We formulate the clinical endpoint prediction task based on EHR data as a representation learning problem of heterogeneous temporal events consists of  asynchronous clinical records from multiple sources.
%This is a general idea (not limited to clinical endpoint prediction)
%and other variants of multi-sequence LSTM could be developed to
%model the multi-source synchronized event-based sequential data ~\cite{neil2016phased}
%in other tasks.

$\bullet$  We propose a novel model called HE-LSTM for learning the representations of heterogeneous event sequence. The model effectively models the multi-scale sampling rates of different kinds of events and their temporal dependency.

%Our proposed model, Multi-Sequence LSTM, equips LSTM with
%carefully designed sequence gates, so that it is not only good
%at modeling the order information in each single sequential data,
%but can also well capture the temporal dependency between
%multiple sequences and handle the multi-scale period of clinical sequences.

$\bullet$ We conducted experiment on real-world clinical data on the tasks of death and abnormal lab tests. Promising results prove the effectiveness of our proposed approach over competitive baselines. 
%show that our method accurately predicts clinical endpoint outcomes such as death and abnormal lab test results.

\section{Related Works}
\subsection{Clinical Endpoint Prediction}
There are plenty of works trying to solve the clinical endpoint prediction problem. However, many of these works only use a small subset of the whole EHR sequences in order to avoid dealing with the high-dimensional event types. Some works select a subset of the clinical events from the EHR data according to the expertise of physicians~\cite{caballero2015dynamically}. For instance,  Alaa
only uses a set of 21 (temporal) physiological streams comprising a set of 11 vital signs and 10 lab test scores to predict ICU admission~\cite{alaa2017learning}.  Some techniques select 50 time series from the whole set of EHR data, and transformed the fixed-size subset into a new latent space using the hyper-parameters of multi-task GP(MTGP) models. They then calculate the similarity of patient's records in the new hyper-parameter space~\cite{ghassemi2015multivariate}. It is notable that manually selecting only a fraction of clinical sequences from original EHR data as the input brings out expert bias, thus these works seldom make full use of the important information of original data.

Most works ignore the content or value of clinical events, and only use the type information of clinical events to predict the endpoints~\cite{Liu2015Temporal}. Specifically,  some approaches train the semantic embeddings for different categories of clinical events for endpoint predictions~\cite{Henriksson2015Modeling}.RETAIN uses two reversed recursive neural networks(RNN) generating attention variables of sequential ICD-9 code groups for the prediction tasks~\cite{choi2016retain}. There are some works using convolution neural network(CNN) to model irregular medical codes for future risk predictions~\cite{nguyen2016deepr}. These works only exploit the type information of historical clinical events to make predictions, ignoring the fine-grained varying attributes of the events. Our work is to address the issue by utilizing the rich type information of clinical events as well as the content and values of the events.

\subsection{Deep Learning Models for Sequential Data}

 Standard RNNs trained with stochastic gradient descent
have difficulty learning long-term dependencies (i.e. spanning
more than 10 time steps) encoded in the input sequences
owing to the vanishing gradient~\cite{hochreiter2001gradient}. The problem has been addressed for example by using a specialized neuron structure in Long Short-Term Memory (LSTM) networks ~\cite{hochreiter1997long} that maintains constant backward flow in the error signal.

In the Clockwork RNN (CW-RNN)~\cite{koutnik2014clockwork}, the hidden layer is partitioned into separate modules, each processing inputs at its own temporal granularity, making computations only at its prescribed clock rate. In this way, the fixed clock periods help to contain long-term dependencies.

Phased LSTM ~\cite{neil2016phased}is a state-of-the-art RNN architecture for modeling event-based sequential data. It extends LSTM by adding the time gate.  The gate has three phases: it rises from 0 to 1 in the first phase and drops from 1 to 0 in the second phase, which are active states. During the third phase, the model is in the inactive state. Updates to $\bm c_t$ and $\bm h_t$ are permitted only in the active state. The Phased LSTM network can achieve fast convergence in most experiments, owing to the fact that the auto-sampling on the long sequential data conducted by the time gate maintains derivative error in the longer back propagation. 

However, these models only focus on learning long-term dependencies in homogeneous sequences, lacking the ability to capture the various and complex temporal dependencies in heterogeneous temporal events, which usually exist in EHR data.

\section{Task Definition}
Here are some notations and the definition of the task.

\subsubsection{Heterogeneous Events Sequence}
The heterogeneous event is defined as the triple $e_i = (type,value,time)$.  $type$ is the category of event, $value$ is the attribute of the event, $type$ and $value$ of $e_i$ are logged at $time$.  %triple sequence
It is noteworthy that the attributes $value$s of different event types  can be either numerical or categorical variable. For example, the attributes of a lab test, e.g.lactate blood  test, is numerical while the attribute of the clinical status, e.g. ectopia type, is categorical variable(i.e. fusion beats, nodal bigeminy).

Heterogeneous events are merged in the ascending order of the record time into a triple sequence $\{e_i\}$. 
We denote the heterogeneous event sequence in a period of time $[T_{start},T_{end}]$ as $\{e_i\}_{T_{start}\leq e_i.t\leq T_{end}}$.
%label sequence

\subsubsection{Clinical Endpoint Prediction Task}

The clinical endpoint prediction task is formulated as follow: given a clinical heterogeneous event sequence $\{e_i\}_{T_{start}\leq e_i.t\leq T_{end}}$, and a binary label $\hat{y}$ for the target endpoint occurring at $ T_{end}$ + $24$ hours, the objective is to predict what the target endpoint is in 24 hours using $\{e_i\}_{T_{start}\leq e_i.t\leq T_{end}}$.

In this paper, we aim to dynamically predict two endpoint outcomes base on the heterogeneous event sequence of patient data in EHR. In the first ``death prediction dataset'', the endpoint outcome is death in either hospital or discharge to home. In the second ``lab test result prediction dataset'', the endpoint outcome is either an abnormal result of the potassium lab test, or clinical stability.

\section{Proposed Method}
%difference compared to LSTM
%Long Short-term Memory (LSTM) networks are good at learning representations of sequences. However, they cannot model sequences which consist of heterogeneous inputs very well, because the single input gate in LSTM lacks the ability to capture the various and complex dependency between different kinds of inputs. On the other hand, LSTM is not sensitive to time information, but time is important in clinical endpoint prediction because a lot of events have their visiting rate and the rate varies significantly across each other. Also, LSTM will be difficult to train when the sequences are too long since the error signals can vanish or explode. Be motivated by these problems, we add an event gate in LSTM which controls the updates of the memory cell. 

%In the meanwhile, since there are less updating, the information in memory cell is more likely to retain, which helps the error signals to flow for a long time span, so that the training will be easier and faster. 

In this section, we introduce the technical details about our proposed model. The overall view of our model is illustrated in Figure 2.

\begin{figure*}[!t]
\centering

\includegraphics[width=.7\textwidth]{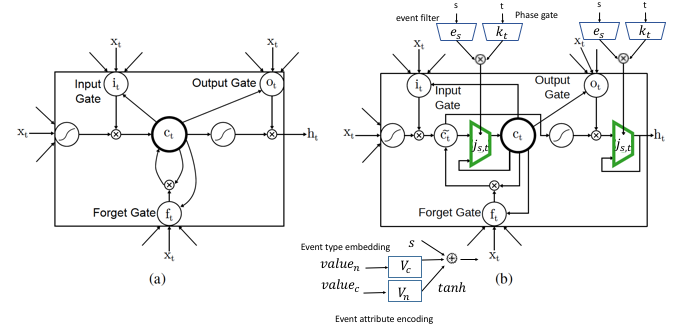}

\caption{Model architecture. (a) Standard LSTM model. (b) HE-LSTM model, with event gate $\bm j_t$ consist of the event filter $\bm e_s$ and phase gate $\bm k_t$  
separately controlled by  the event type $\bm s$ and timestamp $t$. In the HE-LSTM formulation, each neural in the cell value $c_t$ and the hidden output
$h_t$ can be updated during an ``open'' phase by only some certain  types of events; otherwise, the previous values are maintained.}
\label{fig:overview}

\end{figure*}

\subsection{Event type embedding and attribute encoding}
To help the HE-LSTM to trace temporal information of various kinds of events, we use ``event type embedding'' and ``attribute encoding'' to embed the type and attributes of the high dimensional events into compact continuous vectors, which can be trained end-to-end with the following HE-LSTM. 

An event $e_i = (type,value,time)$ of the sequence will be embedded into three parts to feed the HE-LSTM for the endpoint prediction. The three input including  embedding vector of event type $\bm s$, the event attribute encoding vector $\bm x$ and the scale variable time $t$.

The event type vector $\bm s$ carries the information of the event category of $\bm e_i$, and is constructed only by the one hot representation $\bm{type}$ of event type. Similar to word embedding~\cite{mikolov2013distributed}, it will provide a low-dimension vector of the event type with semantic meaning in clinical field. The embedding lookup matrix $C_{type} \in \mathbb{R}^{N \times M}$,  where $N$ is the embedding dimension and $M$ is the number of event types, is established for further training. The event type vector $\bm s$ is given by:

\begin{equation}
\setlength{\abovedisplayskip}{3pt}
\bm s = C_{type} \times \bm{type}
\setlength{\belowdisplayskip}{3pt}
\end{equation}

The event attribute encoding vector $\bm x$ represents the combining information of both event type $\bm{type}$ and the attribute of the event $\bm{value}$, which is the main input of the following HE-LSTM.
Each event has two kinds of attributes. One is categorical with the one hot representation $\bm{value_c} \in \{0,1\}^{C}$, where $C$ is the number of values of categorical attributes of all the event types. The other is numerical with the one hot representation $\bm{value_n}\in \mathbb{R}^{U}$, where $U$ is the number of all numerical attribute  types of all the event types. Notice that the event attribute vector $\bm{value} =  \left[ \bm{value_c},\bm{value_n} \right]$.

Each value of categorical attributes is assigned with a vector from $V_c \in \mathbb{R}^{N \times C}$, where $N$ is the embedding dimension.
As for numerical attributes, they are associated with a value encoding vector in
$V_n \in \mathbb{R}^{N \times U}$, where $N$ is the embedding dimension.

The representing vector of a record $\bm x$ is mainly decided by its event type $\bm s$, however the event attributes  also carries lots of information for modeling patients.
The different values of the same event type, such as the abnormal label in a lab test event, can lead to distinct estimates for the patient's future health status. The other important part of $\bm x$ is a disturbance from the numerical attribute values. For instance, the high numerical value of the lactate blood lab test event indicates potential health problem of the patient, while the low value does not offer much information.
Finally, to combine the three parts of information, the attribute encoding vector  $\bm x$ is given by:

\begin{equation}
\setlength{\abovedisplayskip}{3pt}
\bm x = \bm s + V_c \times \bm{value_c} + \tanh( V_n \times \bm{value_n})
\setlength{\belowdisplayskip}{3pt}
\end{equation}
where   $V_c$ and $V_n$  are parameters to learn.

\subsection{Heterogeneous Event LSTM}

Long short-term memory (LSTM) units~\cite{hochreiter1997long} (Fig. 2(a)) is an important ingredient of modern deep
RNN architectures. We first define their update equations in a commonly-used version in the following:

\begin{align}\label{grubegin}
\setlength{\abovedisplayskip}{3pt}
\bm i_t &= \sigma(W_{ix}\bm x_t+ W_{ih}\bm h_{t-1} + \bm w_{ic} \circ \bm c_{t-1}+ \bm b_i)\\
\bm f_t &= \sigma(W_{fx}\bm x_t+ W_{fh}\bm h_{t-1} + \bm w_{fc} \circ \bm c_{t-1} + \bm b_f)\\
\bm c_t &= \bm f_t \circ \bm c_{t-1} + \bm i_t \circ \tanh\big (W_{cx}\bm x_t+ W_{ch} \bm h_{t-1} + \bm b_c \big) \\
\bm o_t &= \sigma(W_{ox}\bm x_t+ W_{oh}\bm h_{t-1} + \bm w_{oc} \circ \bm c_{t-1} + \bm b_o)\\
\bm h_t &= \bm o_t \circ \tanh \big( \bm c_t \big)
\label{gruend}
\setlength{\belowdisplayskip}{3pt}
\end{align}

%\tilde{\bm h}
%basic LSTM
The main difference from classical RNNs is the use of the gating functions $\bm i_t$, $\bm f_t$, $\bm o_t$, which represent
the input, forget, and output gate at time $t$ respectively. $\bm c_t$ is the cell activation vector, whereas $\bm x_t$
and $\bm h_t$ represent the input feature vector and the hidden output vector respectively. 
The gates use the
typical sigmoid function $\sigma $ and $\tanh$ nonlinear function $\tanh$ with weight parameters
$W_{ih}$, $W_{fh}$, $W_{oh}$, $W_{ix}$, $W_{fx}$, and $W_{ox}$, which connect the different inputs and gates with the memory
cells and outputs, as well as biases $\bm b_i$, $\bm b_f$, and $\bm b_o$. The cell state $\bm c_t$ itself is updated with a fraction of the previous cell state that is controlled by $\bm f_t$, and a new input state created from the element-wise product, denoted by $\circ$, of $\bm i_t$ and the output of the cell state nonlinearity $\tanh$. Optional peephole~\cite{gers2000recurrent} connection weights $\bm w_{ic}$, $\bm w_{fc}$, $\bm w_{oc}$ further influence the operation of the input, forget,
and output gates.

%sequence gate
HE-LSTM extends the LSTM model by adding a new event gate $\bm j_{s,t}$. 
 The event gate has two factors --- an event filter and a phase gate. The event filter only allows the information of a certain cluster of events to fuse into the corresponding memory cell, so that each cell will only trace a particular group of events. Collaborated with the phase gate, the event filter can help the network to maintain the temporal information of the different events in multi-scaled sampling rates. The dependency of the heterogeneous events will be easier to capture by the diverse and long memory of correlated events.
 
 The
opening and closing of this event gate is controlled by the event type embedding $\bm s$ and an independent rhythmic oscillation specified by the phase gate~\cite{neil2016phased} with three parameters. 
And updates to the cell state $\bm c_t$ and $\bm h_t$ are permitted only when the gate is opened.

One factor of the event gate, the event filter $\bm e_{s}$, for each neuron is a feed forward network with a hidden layer of size $L$ with $\tanh$ activation function as following.

\begin{align}
\setlength{\abovedisplayskip}{3pt}
\bm e_s &=  \sigma(W_{em} \tanh\big (W_{ms} \bm s + \bm b_m \big) + \bm b_e)
\setlength{\belowdisplayskip}{3pt}
\end{align}
where $ W_{em} \in \mathbb{R}^{H\times L }$, $ W_{ms} \in \mathbb{R}^{L\times N}$, $\bm b_e\in \mathbb{R}^{H}$ and $\bm b_m\in \mathbb{R}^{L}$ are parameters to learn.

%phased gate
 Considering the multi-scale sampling rates of the events, we extend  the event filter $\bm e_s$ with a time factor proposed in phased LSTM~\cite{neil2016phased} by three parameters:  $t$, $r_{on}$ and $s$, where $t$ represents
the real-time period of the gate, $s$ represents the phase shift and
$r_{on}$ is the ratio of the open phase to the full period. $t$, $r_{on}$ and $s$ are learned by training. Therefore, $\bm j_{s,t}$ is formally defined as:

\begin{align}
\setlength{\abovedisplayskip}{3pt}
\phi_t = \frac{(t-s)\mod \tau}{\tau},
k_t = \begin{cases}
 & \frac{2\phi_t}{r_{on}},\text{ if } \phi_t < \frac{1}{2}r_{on} \\ 
 & 2-\frac{2\phi_t}{r_{on}},\text{ if } \frac{1}{2}r_{on} < \phi_t < r_{on}  \\ 
 & \alpha\phi_t,\text{ otherwise } 
\end{cases}
\setlength{\belowdisplayskip}{3pt}
\end{align}

\begin{equation}
\setlength{\abovedisplayskip}{3pt}
 \bm j_{s,t} = \bm e_s \circ \bm k_t
\setlength{\belowdisplayskip}{3pt}
\end{equation}

%update
Different from traditional RNNs for single sequential data and even sparser variants of RNNs ~\cite{koutnik2014clockwork}, updates in HE-LSTM
can optionally be performed at irregularly sampled time points $t$ for different event types. 
This allows the RNNs to learn the multi-scale rhythm of related events and work with asynchronously sampled
heterogeneous temporal event data. 
We use the shorthand notation $\bm c_l$ = $\bm c_{t_l}$ for cell
states at time $t_l$ (analogously for other gates and units), and let $\bm c_{l-1}$ denote the state at the previous
update time $t_{l-1}$. We can then rewrite the regular LSTM cell update equations for $\bm c_l$ and $\bm h_l$ (from
Eq. 5 and Eq. 7), using proposed cell updates $\tilde{\bm c_l}$ and $\tilde{\bm h_l}$ mediated by the event gate $\bm j_l$ :

\begin{align}
\setlength{\abovedisplayskip}{3pt}
\bm \tilde{c_l} &= \bm f_l \circ \bm c_{l-1} + \bm i_l \circ \tanh\big (W_{cx}\bm x_l+ W_{ch} \bm h_{l-1} + \bm b_c \big) \\
\bm c_l &= \bm j_l \circ \bm \tilde{c_l} + (1-\bm j_l) \circ \bm c_{l-1}\\
\bm \tilde{h_l} &= \bm o_l \circ \tanh \big( \bm \tilde{c_l} \big)\\
\bm h_l &= \bm j_l \circ \bm \tilde{h_l} + (1-\bm j_l) \circ \bm h_{l-1}
\setlength{\belowdisplayskip}{3pt}
\end{align}

%advantage explicit dependency
The HE-LSTM formulation ensures
the flexible allocation and retain of information of each event clusters.
Each neuron of the memory cell and hidden layer of HE-LSTM states can be updated only during the open periods of the event gate. 
In other words, only the information of a certain cluster of events' records can flow into this certain neuron in its own phase.
This is because the event filter $\bm e_s$, one of the factor of the event gate $\bm j_{s,t}$, can be seen as a binary classifier to chose the cluster of event types responsible for each neuron.
Besides, the neuron maintains a perfect memory
during its closed phase, i.e. $c_l = c_{l-\delta} $if $j_{l'} = 0$ for $ t_l \leq  l' \leq t_{l-\delta}$. Thus, other neurons, tracing other events can directly use the information of this cluster of events even they are far away from each other in term of the sequence index. 
Because of this allocation mechanism, HE-LSTM can have much  diverse and longer memory for modeling the dependency of multiple events.

%training

We use a sigmoid layer to predict the true label $\hat{y_t}$ of the  learned representation vector of sequence in the given decision times.

\begin{equation}
\setlength{\abovedisplayskip}{3pt}
y_t = sigmoid(\bm w_p  \bm h_t + b_p )
\setlength{\belowdisplayskip}{3pt}
\end{equation}

where $\bm w_p \in \mathbb{R}^N$ and $b_p$ are parameters to learn.

We use cross-entropy to calculate the classification
loss of the prediction $y_t$ and true label $\hat{y_t}$ of each sample as follows:

\begin{equation}
\setlength{\abovedisplayskip}{3pt}
Loss(\hat{y_t},y_t)
= \frac{1}{N}\sum_{1\leq t \leq N } ( \hat{y_t}\times \ln y_t  +(1-\hat{y_t})\times \ln(1-y_t))
\setlength{\belowdisplayskip}{3pt}
\end{equation}

We can sum up the losses of all the samples in one mini-batch to get the total loss for back propagation.

\section{Experiments}
The source code of MIMIC-III EHR data preprocessing and the proposed model is available and can be found at https://github.com/pkusjh/HELSTM.

\subsection{Data Description and Experimental Settings}
We set up two data sets for evaluation of the models from one real clinical data source. MIMIC-III~\cite{johnson2016mimic}(Medical Information Mart for Intensive Care III) is a large, freely-available database comprising de-identified health-related data relating to over forty thousand patients who stayed in critical care units of the Beth Israel Deaconess Medical Center between 2001 and 2012.

%sequence type
We extract all kinds of events from the MIMIC-III database to get the initial event type  set(18192 kinds of events in total).The statistics of the event types with top frequency are listed in Table.1. By merging the heterogeneous events into triple sequence, we get a set of clinical event sequences. We drop out the sparse event types, whose frequency in total is less than 2500. 

 We extract episodes of patients, which are 24 hours before the occurrence time of each endpoint, from these event sequences as samples. And the upper bound of the record number of the samples is 1000. All the resulting sample events are labeled according to the target endpoint outcome in each task. 

\begin{table}[!t]
\renewcommand\arraystretch{1}
\centering
\begin{tabular}{|l|c|c|}
\hline
\textbf{event sources}        & \textbf{e.g. event types}  & \textbf{\# of types}    \\
\hline

lab test & HEMATOCRIT, & 525 \\
& WHITE BLOOD CELLS&\\
\hline %HEMATOCRIT, WHITE BLOOD CELLS, PLATELET COUNT, PLATELET COUNT, MCH
vital signal      & Heart Rate, & 385\\
&Respiratory Rate&\\
\hline %Heart Rate, Potassium (3.5-5.3), Glucose (70-105), Hematocrit, Respiratory Rate
drug input & 0.9\% Normal Saline, & 60 \\
&Dextrose 5\%&\\
\hline %D5W, 0.9% Normal Saline, Po Intake, NaCl 0.9%, Dextrose 5%
clinical symptom     & Ectopy Type&2382\\
&Motor Response&\\
\hline %Bowel Sounds, Heart Rhythm, Ectopy Type, Eye Opening, Motor Response
procedure     & electrocardiogram & 19 \\
&Invasive Ventilation&\\
\hline %Access Lines - Peripheral, 5-Imaging, Access Lines - Invasive, Invasive Ventilation, EKG
clinical output      & Urine & 17\\
&gastric retentive oral dosage&\\
\hline %Urine Out Foley, Foley, Urine Out Void, Gastric Oral Gastric, OR Urine
\end{tabular}
\caption{Statistics of  the  event type}
\label{tab:some type multiple sequences}
\end{table}

%sequence length
The statistics of the final clinical multiple sequences in two datasets are summarized in Table 2.
\begin{table}[!t]
\renewcommand\arraystretch{1}
\centering
\begin{tabular}{|l|c|c|c|}
\hline
\textbf{Dataset}        & \textbf{\# of samples}  & \textbf{\# of events}    & \textbf{Avg timespan}\\
\hline

death      & 24301(8\%) & 20290879& 3d 15h 58m\\
\hline
lab test & 784583(11\%) & 41006177& 192d 22h 45m \\

\hline
\end{tabular}
\caption{Statistics of  the  datasets(the percentage in the second column is the positive sample rate)}
\label{tab:enropy}
\end{table}

Each dataset is split into 3 parts with fixed proportions, namely training set(70\%), validation set(10\%) and evaluation set(20\%). The data in validation set is used to select hyper-parameters of the proposed and comparing models and to conduct ``early stop'' while training, in which the samples may be different because of cross-validation.
The evaluation set, the details of which is non-transparent for us in the process of training and parameter selection, will then be only used to calculate and report the evaluation metrics for comparison.

\begin{table*}[!t]
\renewcommand\arraystretch{0.9}
\centering

\begin{tabular}{|l||c|c||c|c|}
\hline
\multirow{2}{*}{Methods} & \multicolumn{2}{c||}{death} & \multicolumn{2}{c|}{lab test}  \\
\cline{2-5}            
& \textbf{AUC } 
&  \textbf{AP}
&\textbf{AUC }  
& \textbf{AP }  \\\hline

Independent LSTM      & 0.8771  $\pm$ 0.0005      & 0.5573 $\pm$ 0.0006 & 0.7196 $\pm$ 0.0006& 0.2969 $\pm$ 0.0008 \\
Independent LSTM(shared weight)   & 0.8064  $\pm$ 0.0005       & 0.5301$\pm$ 0.0006 & 0.5308 $\pm$ 0.0005 & 0.1098  $\pm$ 0.0005 \\
\hline
Phased LSTM & 0.8474 $\pm$ 0.0005  & 0.4900 $\pm$ 0.0075& 0.7722 $\pm$ 0.0007& 0.3575 $\pm$ 0.0026 \\
Clock-work RNN       & 0.8400 $\pm$ 0.0001     &  0.7181 $\pm$ 0.0003 & 0.6516 $\pm$ 0.0002 & 0.2208  $\pm$ 0.0003\\

\hline

RETAIN & 0.8967 $\pm$ 0.0011 & 0.5808 $\pm$ 0.0114 & 0.7325 $\pm$ 0.0022& 0.3096 $\pm$ 0.0052\\
LSTM + event embedding \& attr encoding& 0.9466 $\pm$ 0.0002 & 0.7445 $\pm$ 0.0007 & 0.7231 $\pm$ 0.0028& 0.3021 $\pm$ 0.0014\\
\hline

HE-LSTM & \textbf{0.9516  $\pm$ 0.0003}& \textbf{0.7687 $\pm$ 0.0011} & \textbf{0.7987 $\pm$ 0.0008} & \textbf{0.3914 $\pm$ 0.0013} \\

\hline
\end{tabular}

\caption{Performance of clinical endpoint prediction tasks}
\label{tab:result}

\end{table*}

For all
the results presented in this section, the networks, implemented with Theano~\cite{bergstra2010theano},were trained with Adam ~\cite{Kingma2014Adam} set to default learning rate
parameters.

\subsection{Comparing Methods}
We compare HE-LSTM to the following methods.
%no period and no multi sequence

$\bullet$ \textbf{Independent LSTM}
We use LSTM to model each homogeneous event independently and average the resulting representations into a logistic regression layer. Because the computational cost of thousands of independent LSTM exceed our tolerance, we select 25 important events as it was done in many works~\cite{alaa2017learning}.

$\bullet$ \textbf{Independent LSTM (shared weight)}
This model is the same as the previous one, except that the weights in each single LSTM is shared and all events are used as the input of the model.

$\bullet$ \textbf{RETAIN}
RETAIN ~\cite{choi2016retain} mimics physician practice by modeling the EHR data in a reverse time order, and a two-level RNN generating attention variables of sequential data can provides interpretation of the prediction.

%no period and multi-sequence
%$\bullet$ \textbf{Multi-task Gauss Process}

$\bullet$ \textbf{LSTM + event embedding \& attr encoding}
We use the event embedding in the first part of proposed method section as the input of traditional LSTM. Logistic regression is applied to the top hidden layer.

%no multi sequence and period
$\bullet$ \textbf{Clock-work RNN}
Clockwork RNN~\cite{koutnik2014clockwork} described in related works
section.

$\bullet$ \textbf{Phased LSTM}
Phased LSTM~\cite{neil2016phased} described in related works section.

\subsection{Evaluating Metrics}
All methods listed above can produce predict scores instead of binary labels, and the data for target prediction tasks are imbalanced labeled. So metrics for binary labels such as accuracy are not suitable for measuring the performance. Similar to the work~\cite{choi2016retain,Liu2015Temporal}, we adopt the area under ROC curves (Receiver Operating Characteristic curves) and area under PRC (Precision-Recall curves) for evaluation. Both reflect the overall quality of predicted scores at each decision time, according to their true labels. 

$\bullet$ \textbf{the Area under ROC Curve(AUC)}
of comparing $y_i$ with the true label $\hat{y_i}$. AUC is robust to imbalanced positive/negative prediction labels, making it appropriate for evaluating the classification accuracy in the endpoint prediction prediction tasks.

$\bullet$ \textbf{Average Precision(AP)}
Average precision~\cite{turpin2006user} emphasizes ranking positive samples higher. It is the average of precisions computed at the point of each positive samples in the ranked sequence in ascending order of predict score:
\begin{align}
\setlength{\abovedisplayskip}{3pt}
AP = \frac{\sum_{r=1}^N (P(r) \times I(r))}{\mbox{number of positive samples}} \\
P(r) = \frac{|\{\mbox{positive samples of rank r or less}\}|}{r} 
\setlength{\belowdisplayskip}{3pt}
\end{align}
where r is the rank, N the total number of samples, $I()$ a index function on the positive sample of a given rank, and $P(r)$ precision at a given cut-off rank.
 
This metric is also referred to geometrically as the area under the Precision-Recall curve.

$\bullet$ \textbf{Cross Entropy}
that measures the model loss on the test set. The loss can be calculated
by Eq (16).

\subsection{Quantitative Results}

\begin{table*}[!t]
\renewcommand\arraystretch{0.7}
\centering

\begin{tabularx}{0.9\linewidth}{|c|>{\centering\arraybackslash}X|c|c|c|}
\hline
\multicolumn{2}{|c|}{\textbf{Methods}  }          
& \textbf{Phase gate} 
&\textbf{Event filter }  
&\textbf{Event gate }  
 \\\hline
\multirow{6}{*}{\begin{tabular}[c]{@{}l@{}}death \\ prediction\end{tabular}}
&AUC(1st epoch) & 0.9301   & 0.9105 & \textbf{0.9370} \\
&AUC   & 0.9471         & \textbf{0.9518} & 0.9516  \\
&AP(1st epoch)     & 0.6856      & 0.6048 & \textbf{0.7094} \\
&AP    & 0.7467        & 0.7679  & \textbf{0.7687}\\
&Entropy(1st epoch)     & 0.1561        & 0.1835 & \textbf{0.1479}\\
&Entropy    & 0.1369        & 0.1301  & \textbf{0.1297}\\

\hline
\multirow{6}{*}{\begin{tabular}[c]{@{}l@{}}abnormal lab test \\ prediction\end{tabular}} &AUC(1st epoch) & 0.7050   & 0.6747 & \textbf{0.7275} \\
&AUC   & 0.7945          & 0.7559  & \textbf{0.7987} \\
&AP(1st epoch)     & 0.2752         & 0.2403  &\textbf{0.2965 }\\
&AP    & 0.3875         & 0.3410  & \textbf{0.3914}\\
&Entropy(1st epoch)     & 0.3373         & 0.3448  & \textbf{0.3298}\\
&Entropy    & 0.3019         & 0.3178  & \textbf{0.3003}\\
\hline
\end{tabularx}

\caption{Performance with different settings of the event gate}
\label{tab:variation model result}

\end{table*}

Table.3 shows the area under ROC and AP of different methods on death and lab test datasets respectively.  From the results in Table.3, we draw the following conclusions:

%dependency of multiple sequences
Firstly, models considering the dependency of correlated events types outperform all the independent sequential models and the proposed HE-LSTM achieves the best performance. For example,  on death prediction task,  RETAIN, LSTM and HE-LSTM improve the AP of lab test prediction by around 4.3\%, 2.4\% and 32.1\% respectively compared to the best of ``independent LSTM'' model without weight share of the parameters in each independent LSTMs. The similar results have been shown in other experiments and metrics. Furthermore, our model achieves the highest performance among these heterogeneous sequential models. For example, on lab test prediction task, HE-LSTM improves the AP by 26.2\% and 29.4\% compared to  RETAIN and LSTM. Besides, the improvements on AUC are 9.0\% and 10.4\% respectively. We can draw the conclusion that the dependency information of correlated clinical temporal events is useful in endpoint prediction tasks and learning joint representations is more effective to model the temporal dependency of different events of EHR data compared to simple independent sequential models.

%adaptively fit the multiscaled sampling rates
Secondly, compared to the densely updating recurrent neural networks, the RNNs adaptive to the sampling rate pattern of events make more improvement of the prediction performance. 
For example, clock-work RNN  improve the AP of death prediction by 29.0\% and 33.9\% compared to the two kinds of independent LSTMs, while the improvements of AUC and AP are 7.0\% and 20.7\% for phased LSTM compared to the best of independent LSTMs in the lab test prediction task.
We can draw the conclusion that multi-scaled sampling rate pattern of events is effective for endpoint prediction, which makes the model concentrate more on the important events in different phases other than treating all clinical events equally in the long sequence.

%joint consideration of dependency of multi-sequence and periodical pattern
Thirdly,  HE-LSTM achieves the best performance on all datasets and all evaluation metrics. HE-LSTM outperforms all sparsely updating recurrent neural networks and heterogeneous sequential models on each metrics of two datasets. 
Models solely utilizing multi-scale sampling pattern in event sequence or models straight-forwardly merging different type of events are not the best choice for clinical endpoint prediction in EHR data. 
Take the result of death prediction for example, HE-LSTM improves the AUC and AP by 12.4\% and 7.0\% respectively compared to the best of sparsely updating  methods without the event type embedding and attribute encoding modules. 
The improvements of HE-LSTM compared to the heterogeneous sequential models without event gates are 3.4\% and 30.3\% in average in term of AUC and AP. 
We can draw the conclusion that the proposed HE-LSTM effectively improve the performance because of the joint effects of tracing the temporal dependency of heterogeneous events and adaptively fitting their multi-scaled sampling patterns.

\subsubsection{Experiment on variations of the event gate}
%objective
To evaluate the effect of the components in the event gate $\bm j_{s,t}$, we replace  $\bm j_{s,t}$ in (Eq 10) with its factors, namely phase gate $\bm k_t$ and event filter $\bm e_s$ while remaining the other parts of the model identical.
%what experiment
The results on two datasets are list in table 4, including AUC, AP and cross entropy on test data as well as the values of the three metrics when the first training epoch is finished.  

%basic result and e.g.
The event filter mainly helps to improve the performance of clinical endpoint prediction tasks by modeling the dependency of heterogeneous events. Both the event gate and the event filter achieve good performance in all metrics and both datasets when the training is finished.
For example, the event gate and the event filter improve the AUC of death prediction by 0.5\% and 0.5\% compared to the phase gate, while the improvements of AP are 2.8\% and 2.9\% and the improvements of entropy are 4.9\% and 5.2\%.

The phase gate helps to achieve a fast convergence in the early stage of training by fitting the multi-scaled sampling rates of different events. HE-LSTM and the model with only phase gate get much higher performance in all metrics and both datasets in the first epoch of training.
Take results in lab test task for example, the phase gate and the event gate improve the AUC in first epoch by 4.6\% and 7.9\% compared to the event filter, while the improvements of AP are 14.5\% and 23.3\% and the improvements of entropy are 2.2\% and 4.3\%.

%conclusion
From these comparisons, we draw the conclusion that the event filter and the phase gate collaborates jointly in modeling the dependency in heterogeneous temporal events with the multi-scale sampling rates, which leads to the accurate and efficient performance on the clinical endpoint prediction task.

\subsubsection{Experiment on varying length of multiple sequential data}

To evaluate the ability to model the temporal dependency of heterogeneous temporal events of our proposed architect and the other baselines, we feed the trained models multiple events in test set with various length, in the range of 20 to 1000, as input. From figure 3, we can draw the following conclusions:

Firstly, temporal information is effective for endpoint prediction tasks. The performances of most models improve with the increase of the input sequence length. Especially, the performance increases sharply when the length of input sequence is less than 200.

Secondly, HE-LSTM is better at handling the dependency of heterogeneous temporal events than other models.
When the input sequence is short, the performances of different models are similar. The reason lies in the fact that, for short sequence input, the combination of independent representations of a single event makes less difference from the joint representation of heterogeneous events in HE-LSTM.
But when the input sequence get longer and longer, the performance of our model steadily increase from 0.7551 to 0.7687 in term of AP and from 0.9482 to 0.9516 in term of AUC. The performance of other models remained almost unchanged at almost 0.9465 of AUC and 0.7434 of AP.

\begin{figure}[!t]
\centering
\includegraphics[width=1.1\linewidth]{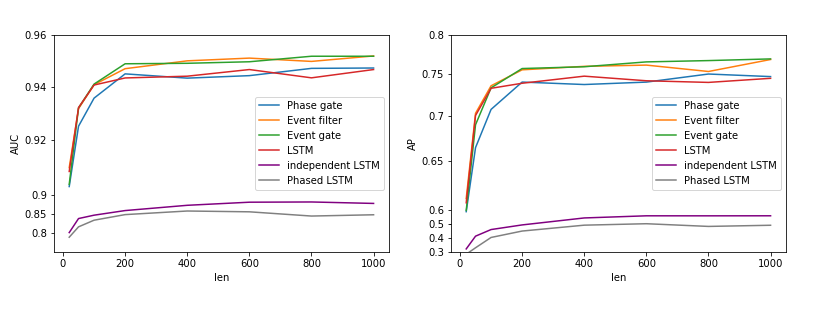}

\caption{the performance on death prediction task with varying input length of the heterogeneous event sequence data }

\label{fig:length}
\end{figure}

%\subsubsection{Event type embeddings clusters}
%objective
%what experiment
%basic result and e.g.
%conclusion

%later
%\subsubsection{The transform matrix for dependency of event clusters}

\subsubsection{Different initial period}

To explore the effect of the event filter in the event gate when modeling heterogeneous sequential EHR data, we compare the performance of the proposed HE-LSTM with the reduced HE-LSTM, of which the event filter factor in the event gate is removed. 
We use different initial periods of $\tau$ during training for death prediction task.
The period  was drawn uniformly in the exponential
domain, comparing four sampling intervals $\exp(U(1, 2))$, $\exp(U(2, 3))$, $\exp(U(3, 4))$, and
$\exp(U(4, 5))$ for each model. 
The results in Figure.4 show that the initialization of $\tau$ affects the performance of both models. But HE-LSTM is more robust to the initialization. For example, the improvements of HE-LSTM compared to the one without event filter are 4.1\%, 4.1\%, 2.8\%and 6.6\% on average.
We can draw the conclusion that, with the help of event filter, the event gate can be more adaptive to multi-scale sampling rates of the events in the heterogeneous temporal sequence. 

\begin{figure}[!t]
\centering
\includegraphics[width=1.09\linewidth]{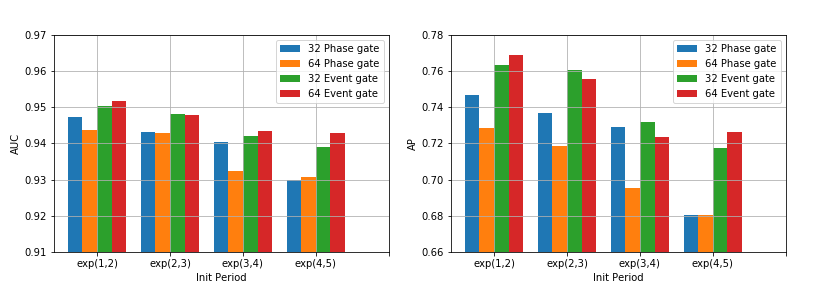}

\caption{Different initial period}

\label{fig:period}
\end{figure}

%\subsubsection{Learned gate attention and rhythms and compare with the data}

%\subsubsection{Fast convergence}

\section{Conclusion}
In this paper, we propose a novel HE-LSTM model
to learn joint representations of heterogeneous temporal events for clinical endpoint prediction.
Our model can adaptively fit the multi-scaled sampling rates of events in the heterogeneous event sequence. By tracing the temporal information of different kinds of events in the long sequence, the temporal dependency of different types of events can be captured in our learned representations. 
Experimental results with real-world clinical data on the tasks of predicting death and abnormal lab tests prove the effectiveness of our proposed approach over competitive baselines. 

\section{Acknowledgement}
This paper is partially supported by the National Natural Science Foundation of China (NSFC Grant Nos.91646202, 61772039 and 61472006).

\bibliography{aaai}

\begin{thebibliography}{}

\bibitem[\protect\citeauthoryear{Alaa, Hu, and van~der
  Schaar}{2017}]{alaa2017learning}
Alaa, A.~M.; Hu, S.; and van~der Schaar, M.
\newblock 2017.
\newblock Learning from clinical judgments: Semi-markov-modulated marked hawkes
  processes for risk prognosis.
\newblock {\em arXiv preprint arXiv:1705.05267}.

\bibitem[\protect\citeauthoryear{Bergstra \bgroup et al\mbox.\egroup
  }{2010}]{bergstra2010theano}
Bergstra, J.; Breuleux, O.; Bastien, F.; Lamblin, P.; Pascanu, R.; Desjardins,
  G.; Turian, J.; Warde-Farley, D.; and Bengio, Y.
\newblock 2010.
\newblock Theano: A cpu and gpu math compiler in python.
\newblock In {\em Proc. 9th Python in Science Conf},  1--7.

\bibitem[\protect\citeauthoryear{Blunsom \bgroup et al\mbox.\egroup
  }{2017}]{blunsom2017characters}
Blunsom, P.; Cho, K.; Dyer, C.; and Schütze, H.
\newblock 2017.
\newblock From characters to understanding natural language (c2nlu): Robust
  end-to-end deep learning for nlp (dagstuhl seminar 17042).
\newblock In {\em Dagstuhl Reports}, volume~7.
\newblock Schloss Dagstuhl-Leibniz-Zentrum fuer Informatik.

\bibitem[\protect\citeauthoryear{Caballero~Barajas and
  Akella}{2015}]{caballero2015dynamically}
Caballero~Barajas, K.~L., and Akella, R.
\newblock 2015.
\newblock Dynamically modeling patient's health state from electronic medical
  records: A time series approach.
\newblock In {\em Proceedings of the 21th ACM SIGKDD International Conference
  on Knowledge Discovery and Data Mining},  69--78.
\newblock ACM.

\bibitem[\protect\citeauthoryear{Choi \bgroup et al\mbox.\egroup
  }{2016}]{choi2016retain}
Choi, E.; Bahadori, M.~T.; Sun, J.; Kulas, J.; Schuetz, A.; and Stewart, W.
\newblock 2016.
\newblock Retain: An interpretable predictive model for healthcare using
  reverse time attention mechanism.
\newblock In {\em Advances in Neural Information Processing Systems},
  3504--3512.

\bibitem[\protect\citeauthoryear{Gers and
  Schmidhuber}{2000}]{gers2000recurrent}
Gers, F.~A., and Schmidhuber, J.
\newblock 2000.
\newblock Recurrent nets that time and count.
\newblock In {\em Neural Networks, 2000. IJCNN 2000, Proceedings of the
  IEEE-INNS-ENNS International Joint Conference on}, volume~3,  189--194.
\newblock IEEE.

\bibitem[\protect\citeauthoryear{Ghassemi \bgroup et al\mbox.\egroup
  }{2015}]{ghassemi2015multivariate}
Ghassemi, M.; Pimentel, M.~A.; Naumann, T.; Brennan, T.; Clifton, D.~A.;
  Szolovits, P.; and Feng, M.
\newblock 2015.
\newblock A multivariate timeseries modeling approach to severity of illness
  assessment and forecasting in icu with sparse, heterogeneous clinical data.
\newblock In {\em Proceedings of the AAAI Conference on Artificial
  Intelligence. AAAI Conference on Artificial Intelligence}, volume 2015,  446.
\newblock NIH Public Access.

\bibitem[\protect\citeauthoryear{He \bgroup et al\mbox.\egroup
  }{2016}]{he2016deep}
He, K.; Zhang, X.; Ren, S.; and Sun, J.
\newblock 2016.
\newblock Deep residual learning for image recognition.
\newblock In {\em Proceedings of the IEEE conference on computer vision and
  pattern recognition},  770--778.

\bibitem[\protect\citeauthoryear{Henriksson \bgroup et al\mbox.\egroup
  }{2015}]{Henriksson2015Modeling}
Henriksson, A.; Zhao, J.; Bostrom, H.; and Dalianis, H.
\newblock 2015.
\newblock Modeling electronic health records in ensembles of semantic spaces
  for adverse drug event detection.
\newblock In {\em IEEE International Conference on Bioinformatics and
  Biomedicine},  343--350.

\bibitem[\protect\citeauthoryear{Hinton \bgroup et al\mbox.\egroup
  }{2012}]{hinton2012deep}
Hinton, G.; Deng, L.; Yu, D.; Dahl, G.~E.; Mohamed, A.-r.; Jaitly, N.; Senior,
  A.; Vanhoucke, V.; Nguyen, P.; Sainath, T.~N.; et~al.
\newblock 2012.
\newblock Deep neural networks for acoustic modeling in speech recognition: The
  shared views of four research groups.
\newblock {\em IEEE Signal Processing Magazine} 29(6):82--97.

\bibitem[\protect\citeauthoryear{Hochreiter and
  Schmidhuber}{1997}]{hochreiter1997long}
Hochreiter, S., and Schmidhuber, J.
\newblock 1997.
\newblock Long short-term memory.
\newblock {\em Neural computation} 9(8):1735--1780.

\bibitem[\protect\citeauthoryear{Hochreiter \bgroup et al\mbox.\egroup
  }{2001}]{hochreiter2001gradient}
Hochreiter, S.; Bengio, Y.; Frasconi, P.; Schmidhuber, J.; et~al.
\newblock 2001.
\newblock Gradient flow in recurrent nets: the difficulty of learning long-term
  dependencies.

\bibitem[\protect\citeauthoryear{Johnson \bgroup et al\mbox.\egroup
  }{2016}]{johnson2016mimic}
Johnson, A.~E.; Pollard, T.~J.; Shen, L.; Lehman, L.-w.~H.; Feng, M.; Ghassemi,
  M.; Moody, B.; Szolovits, P.; Celi, L.~A.; and Mark, R.~G.
\newblock 2016.
\newblock Mimic-iii, a freely accessible critical care database.
\newblock {\em Scientific data} 3.

\bibitem[\protect\citeauthoryear{Kingma and Ba}{2014}]{Kingma2014Adam}
Kingma, D., and Ba, J.
\newblock 2014.
\newblock Adam: A method for stochastic optimization.
\newblock {\em Computer Science}.

\bibitem[\protect\citeauthoryear{Koutnik \bgroup et al\mbox.\egroup
  }{2014}]{koutnik2014clockwork}
Koutnik, J.; Greff, K.; Gomez, F.; and Schmidhuber, J.
\newblock 2014.
\newblock A clockwork rnn.
\newblock In {\em International Conference on Machine Learning},  1863--1871.

\bibitem[\protect\citeauthoryear{Liu \bgroup et al\mbox.\egroup
  }{2015}]{Liu2015Temporal}
Liu, C.; Wang, F.; Hu, J.; and Xiong, H.
\newblock 2015.
\newblock Temporal phenotyping from longitudinal electronic health records: A
  graph based framework.
\newblock In {\em ACM SIGKDD International Conference on Knowledge Discovery
  and Data Mining},  705--714.

\bibitem[\protect\citeauthoryear{Mikolov \bgroup et al\mbox.\egroup
  }{2010}]{mikolov2010recurrent}
Mikolov, T.; Karafi{\'a}t, M.; Burget, L.; Cernock{\`y}, J.; and Khudanpur, S.
\newblock 2010.
\newblock Recurrent neural network based language model.
\newblock In {\em Interspeech}, volume~2, ~3.

\bibitem[\protect\citeauthoryear{Mikolov \bgroup et al\mbox.\egroup
  }{2013}]{mikolov2013distributed}
Mikolov, T.; Sutskever, I.; Chen, K.; Corrado, G.~S.; and Dean, J.
\newblock 2013.
\newblock Distributed representations of words and phrases and their
  compositionality.
\newblock In {\em Advances in neural information processing systems},
  3111--3119.

\bibitem[\protect\citeauthoryear{Neil, Pfeiffer, and
  Liu}{2016}]{neil2016phased}
Neil, D.; Pfeiffer, M.; and Liu, S.-C.
\newblock 2016.
\newblock Phased lstm: Accelerating recurrent network training for long or
  event-based sequences.
\newblock In {\em Advances in Neural Information Processing Systems},
  3882--3890.

\bibitem[\protect\citeauthoryear{Nguyen \bgroup et al\mbox.\egroup
  }{2016}]{nguyen2016deepr}
Nguyen, P.; Tran, T.; Wickramasinghe, N.; and Venkatesh, S.
\newblock 2016.
\newblock Deepr: A convolutional net for medical records.
\newblock {\em IEEE Journal of Biomedical and Health Informatics}.

\bibitem[\protect\citeauthoryear{Turpin and Scholer}{2006}]{turpin2006user}
Turpin, A., and Scholer, F.
\newblock 2006.
\newblock User performance versus precision measures for simple search tasks.
\newblock In {\em Proceedings of the 29th annual international ACM SIGIR
  conference on Research and development in information retrieval},  11--18.
\newblock ACM.

\end{thebibliography}
\bibliographystyle{aaai}

\end{document}